%% file: 00_main.tex
\documentclass[letterpaper, 10pt, conference]{ieeeconf}      

\IEEEoverridecommandlockouts                              
\usepackage[utf8]{inputenc}
\usepackage[english]{babel}

\usepackage{cite}
\usepackage{multirow}
\usepackage{graphicx} 
\usepackage[table,xcdraw]{xcolor}
\usepackage{epsfig} 

\usepackage{amsmath} 
\usepackage{amssymb}  
\usepackage{amsthm}

\theoremstyle{plain}

\usepackage{flafter} 

\usepackage{caption}
\usepackage{subcaption}

\usepackage{algorithm}
\usepackage[]{algpseudocode} 

\usepackage{program}

\usepackage{breqn}

\usepackage{siunitx} 
\usepackage{tabularx}

\usepackage{hyperref}
\usepackage{cleveref}

\newcolumntype{L}[1]{>{\raggedright\arraybackslash}p{#1}}
\newcolumntype{C}[1]{>{\centering\arraybackslash}p{#1}}
\newcolumntype{R}[1]{>{\raggedleft\arraybackslash}p{#1}}

\usepackage[nolist,nohyperlinks]{acronym}

\graphicspath{{figures/}}

\title{\LARGE \bf RoBoa: Construction and Evaluation of a Steerable Vine Robot for Search and Rescue Applications}%

\author{Pascal Auf der Maur$^{*1}$, Betim Djambazi$^{*1}$, Yves Haberthür$^{*1}$,  Patricia Hörmann$^{*1}$, Alexander Kübler$^{*1}$,\\ Michael Lustenberger$^{*1}$, Samuel Sigrist$^{*1}$, Oda Vigen$^{*1}$, Julian F\"orster$^{1}$, Florian Achermann$^{1}$, Elias Hampp$^{1}$,\\ Robert K. Katzschmann$^{2}$, and Roland Siegwart$^{1}$ 
\thanks{$^{*}$ Equal contribution.}
\thanks{$^{1}$ Autonomous Systems Lab, ETH Zürich, Switzerland.}
\thanks{$^{2}$ Soft Robotics Lab, ETH Zürich, Switzerland.}
\thanks{Corresponding author: Pascal Auf der Maur ({\tt\small pascalau@ethz.ch}).}}

\begin{document}

\maketitle
\thispagestyle{empty}
\pagestyle{empty}


\begin{abstract}
\input{00a_abstract}
\end{abstract}


\input{01_introduction}

\input{02_design}

\input{03_results}

\input{04_conclusion}








\section*{ACKNOWLEDGMENT}
We are grateful for the support of the Swiss Drone and Robotics Centre, the Swiss Rescue Troops, our sponsors who gave us the opportunity to develop our prototype, and the helpful advice and assistance of Michael Riner-Kuhn, Matthias Müller, Cornelia Della Casa, Luciana Borsatti, and Nicholas Lawrance.


\bibliographystyle{IEEEtran}
\bibliography{IEEEabrv,05_references}

\begin{acronym}
\acro{imu}[IMU]{Inertial Measurement Unit}
\acro{sar}[SaR]{Search and Rescue}
\acro{poe}[PoE]{Power over Ethernet}
\acro{sbc}[SBC]{Single Board Computer}
\acro{pcb}[PCB]{Printed Circuit Board}
\acro{slam}[SLAM]{Simultaneous Localization and Mapping}
\end{acronym}

\end{document}

%% file: 00a_abstract.tex
RoBoa is a vine-like search and rescue robot that can explore narrow and cluttered environments such as destroyed buildings. The robot assists rescue teams in finding and communicating with trapped people. It employs the principle of vine robots for locomotion, everting the tip of its tube to move forward. Inside the tube, pneumatic actuators enable lateral movement. The head carries sensors and is mounted outside at the tip of the tube. At the back, a supply box contains the rolled up tube and provides pressurized air, power, computation, as well as an interface for the user to interact with the system. A decentralized control scheme was implemented that reduces the required number of cables and takes care of the low-level control of the pneumatic actuators. The design, characterization, and experimental evaluation of the system and its crucial components is shown. The complete prototype is fully functional and was evaluated in a realistic environment of a collapsed building where the remote-controlled robot was able to repeatedly locate a trapped person after a travel distance of about \SI{10}{\metre}.

%% file: 01_introduction.tex
\section{Introduction} \label{sec:introduction}

%
\subsection{Motivation}
Whenever natural catastrophes such as earthquakes or landslides strike near populated areas, chances are high that people get buried in destroyed buildings. If not found, rescued, and cared for within 72 hours, the chances of survival are low \cite{10.1145/1582379.1582514}.

Today, rescue workers are mostly limited to rudimentary tools and techniques. In a typical scenario rescue forces first try to identify the rough location of potentially trapped people using rescue dogs. Subsequently, a squad assembles at the site and tries to contact victims by alternately knocking and listening for any answers. The success of this method depends on whether the victim is conscious or not. Consequently, people requiring rescue most urgently cannot be located with this method. Since the debris is structurally highly unstable, the rescue forces risk their lives having little information about the dangers they are about to face.

\begin{figure}
    \centering
    \includegraphics[width=0.48
    \textwidth]{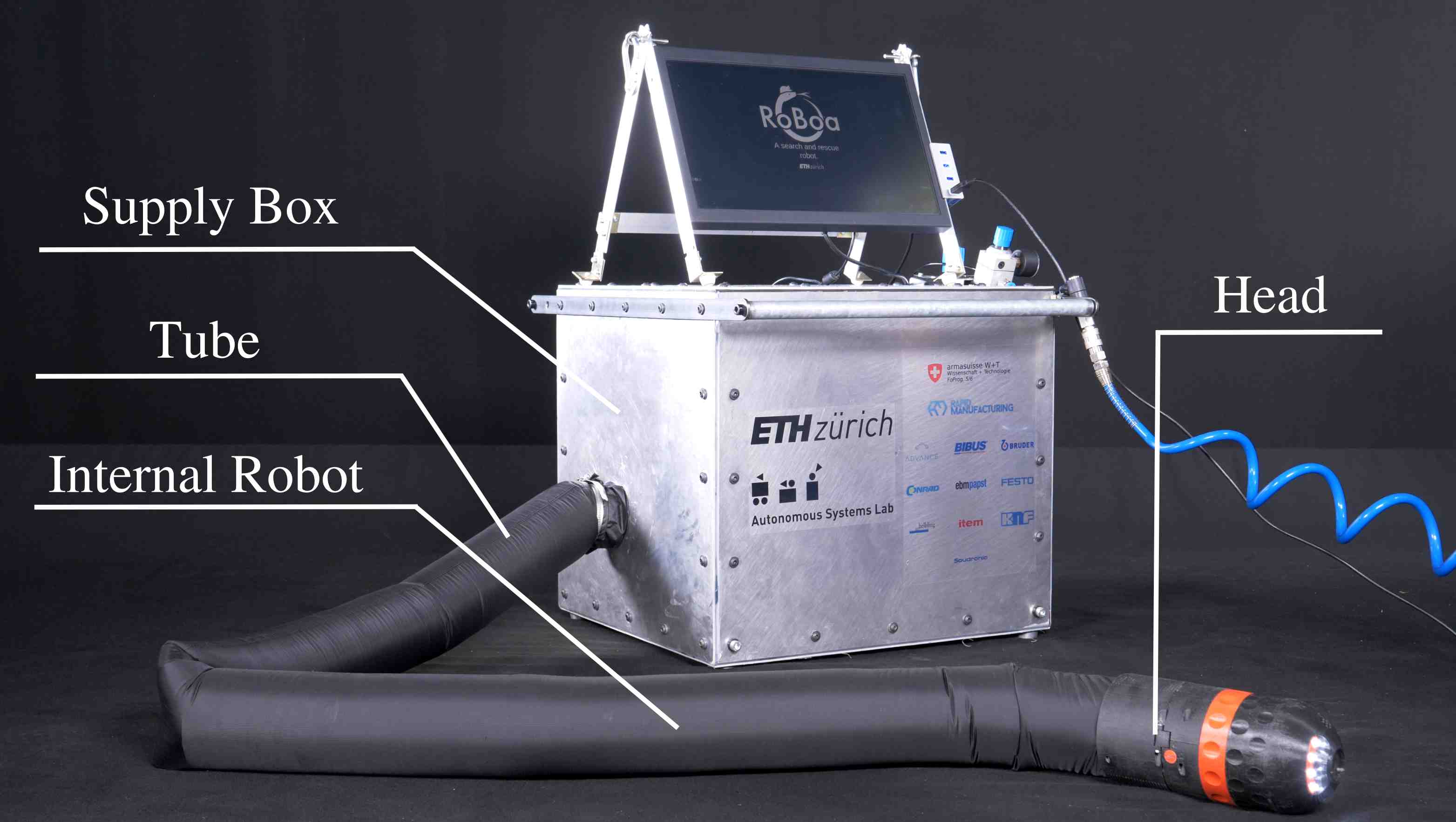}
    \caption{The RoBoa system consists of the supply box from which the everting robot operates. The components of the everting robot are an inflatable fabric tube with an internal steerable robot, and a head with a camera and sensors. The internal robot is hidden inside the tube, right behind the head.}
    \label{fig:overview}
\end{figure}

To overcome these challenges, we propose the vine-like \ac{sar} robot \textit{RoBoa} shown in Fig.\,\ref{fig:overview}. The robot is able to maneuver in three-dimensional rubble fields, allowing rescue workers to find buried victims in a safer, faster and more reliable way by using a unique combination of locomotion and steering.

\subsection{Related Work}

Several different kinds of \ac{sar} robots can be found in literature. Drones can detect people and give an overview of the whole situation by mapping the area\,\cite{millane2018c} or detecting people on the surface\,\cite{drone_stateof}, but they cannot get into the debris to find buried victims. Track-based robots can easily cross rough terrain on the surface but have trouble getting deeper into the debris since they need to be connected to a base either by a cable or wirelessly\,\cite{yamauchi2004packbot, soryu, inachus}. A cable introduces considerable amounts of friction while wireless signals cannot pass through the iron reinforcements of the debris. The same issue occurs with wheel-based\,\cite{6386291}, or legged robots\,\cite{7758092, cockroachrobot}. Worm-like robots can be smaller than wheel-based robots but they still need to be connected to a base by a cable or wirelessly\,\cite{activescope}.

In contrast, growing robots have the advantage that the outside hull remains static with respect to the environment. Therefore, the friction to the surroundings is insignificant when moving. Growing can for example be achieved by 3D printing \cite{sadeghi2017toward}. However this has difficulties with overheating. After \SI{15}{\minute} the heat transfers to other parts of the system which causes problems in electronic parts and leads to softening of the unused filament. The system is also rather slow with a maximal growing speed of \SI{4}{\milli\metre\per\minute}.

Another approach to growing is taken by vine robots. Vine robots do not have to overcome the friction from the outside hull to the environment but only the internal friction within the robot, which is simpler to control. These robots thus have the capability to carry cables to the tip by controlling the friction inside of the robot.
Hawkes et al.\,\cite{vinepaper} describes an everting tube, which is characteristic of vine robots,
and provides application ideas such as acting as a fire hose or a pneumatic jack, exploring new places with a camera at the tip or the usage as a \ac{sar} robot.
Another variant of vine robots consists of multiple smaller tubes\,\cite{Tsukagoshi2011TipGA}.
Current methods for steering vine robots attach a soft structure such as series pouch motors to the whole body of the vine robot\,\cite{7989648, doi:10.1089/soro.2018.0034, Coad_2020, niiyama2015pouch}, in the following referred to as \emph{body-steered}. Because the pouch motors are present along the full length of the robot's tube, manufacturing such a vine robot becomes more difficult the longer it gets. Jeong et al.\,\cite{jeong2020tip} show a retraction mechanism for vine robots by pulling the material back right at the tip in order to avoid kinks. Different methods for mounting sensors to the tip of the vine robot have been explored in literature, a form fit\,\cite{jeong2020tip} is preferred over a magnetic attachment\,\cite{luong2019eversion}, since the latter detaches more easily in rough environments.

The steering mechanism of the robot described in this paper includes 3D-printed pneumatic actuators that consist of three strands that elongate when put under pressure. This actuator principle is based on\,\cite{Hadzic2018, festoproboscis}.

\subsection{Contributions}
\begin{itemize}
    \item A new steering concept for vine robots: By putting pneumatic actuators, which always stay at the front, inside the vine robot, the robot is able to change its shape and thus steer through debris. In comparison with a previous mechanism design\,\cite{7989648}, the bending radius is reduced to \SI{20}{\centi\metre}.
    \item The longest, steerable vine robot (to the best of our knowledge): Being able to steer when fully extended at \SI{17}{\metre} length, the robot outgrows previous steerable robots\,\cite{7989648} by \SI{70}{\%}. Previously, the whole tube had to change its shape, and thus steering and manufacturing the tube becomes challenging. The robot described here only steers the tip, making the steering mechanism independent of the length and simplifying manufacturing. 
    \item Resistance against adverse conditions encountered in real-world environments: The robot was not only tested in a laboratory environment but also multiple times in the debris of a destroyed house.
    \item A moving pressure supply: The valve terminal is mounted onto the pneumatic actuators. Those actuators always stay at the tip of the robot and thus the valve terminals move with the speed of the robot's tip. Thus only one pressure tube has to be carried to the tip.
\end{itemize}

%% file: 02_design.tex
\section{Design}
\label{sec:design}

In this section we present the design of our robot and the decisions made during its construction. An overview of the system is shown in Fig.\,\ref{fig:overview}.

\subsection{Requirements}
\label{requirements}

The main goal of this work was to develop a concept for a \ac{sar} robot that supports rescue workers in their missions.
Thanks to the collaboration with the Swiss Rescue Troops, the requirements for the use in a \ac{sar} environment were determined (e.g. fields of debris after an earthquake):


\begin{enumerate}
    \item The robot has to operate under debris and perform sustained operations in this adverse environment. It has to deal with dust, darkness, small holes and spaces, and obstacles like stones or collapsed structures. Thus, the \emph{locomotion} principle of the robot must minimize friction within such environments and navigate through tight spaces. The locomotion principle should also enable backwards movement.
    \item According to the Swiss Rescue Troops' experience, the system would be advantageous if it is longer than \SI{3}{\meter}. At a \emph{length} of \SI{15}{\metre} or more almost every victim could be reached. Therefore, it was decided to set this length as the required length of the robot.
    \item Because the Swiss Rescue Troops use drills with a diameter of up to \SI{112}{\milli\meter} to enter destroyed buildings, the maximum \emph{diameter} of the robot is constrained by this value.
    \item To allow for sufficient \emph{maneuverability} in debris, the robot has to be bendable 90$^\circ$ horizontally as well as 45$^\circ$ vertically at a turning radius of \SI{25}{\centi\metre}.
    \item Different sensors need to be implemented at the tip of the robot to allow \emph{localization} of victims under challenging lighting conditions and enable audio \emph{communication}.
    \item The robot can be \emph{operated} by a trained user.
\end{enumerate}

The work towards the final prototype was then guided by those requirements to realize a system that addresses the needs of rescue workers as completely as possible. \Cref{sec:results} discusses how well the requirements were met.




%
\subsection{Everting Tube}

\begin{figure}
    \centering
    \includegraphics[width=0.48\textwidth]{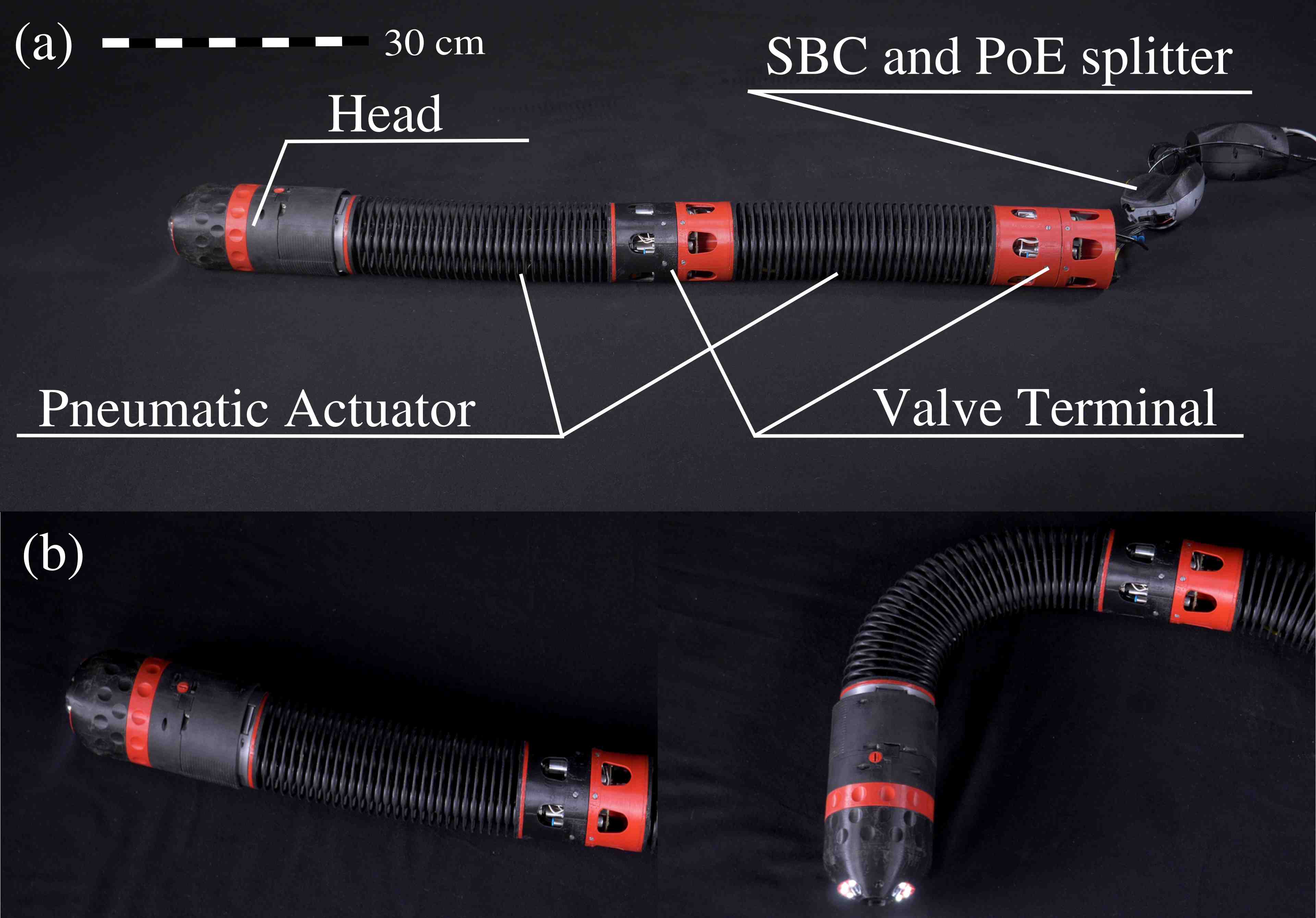}
     \caption{(a) The internal robot with the attached head is shown. Highlighted are the sensing head, the pneumatic actuator with valve terminal for lateral movements, and the cases containing the \ac{sbc} and \ac{poe} splitter. (b) Maximum achievable deflection of a single actuator segment without the tube.}
    \label{fig:IR}
\end{figure}

As a first step during the development, different propulsion and robot types were analyzed for their suitability in the aforementioned environment. This choice was strongly influenced by the need for communication between the robot's tip (in the debris) and its base station (outside of the debris). For this, we tested a wireless connection with an agent located under the debris using a miniaturized computer (Raspberry Pi 4). During the experiment, a stable connection could only
be achieved up to a distance of \SI{5}{\metre} with conventional WiFi. Since a requirement is operation with lengths of up to \SI{15}{\metre}, the only viable alternative is a tether between tip and base station. However, this raises the problem of friction between the environment and the cable being dragged along by the tip. Experience from the predecessor project called Proboscis\,\cite{Hadzic2018} showed that tracked robots with a diameter of \SI{112}{\milli\metre} (as per the requirements) have difficulties overcoming this friction.

In order to control the friction between cables and environment, we opted for locomotion based on the eversion principle used by vine robots\,\cite{vinepaper}. The main component is a furled tube made of a \SI{20}{Denier} coated ripstop nylon with welded seams that is everted by pressurizing it with air. The tube material exposed to the outside (referred to as \emph{outer tube}) does not move forward relative to the environment, but is elongated by transporting more tube material on the inside (referred to as \emph{inner tube}) to the front and everting it there. The decisive advantage of this approach is that the robot has to overcome hardly any friction with the rough and unpredictable surroundings. For a detailed introspection on how our design interacts with the tube see Fig.\,\ref{fig:tube}.

However, the friction does not disappear, but instead occurs on the inside of the tube. Specifically, the friction comes about due to velocity differences between inner and outer tube, as well as between inner tube, internal robot, and cables. Since the inside is fully subject to our design, it is easier to control friction there. For example, the smoothness of the surface of the tube can be influenced through a favourable material pairing or the use of lubricants.

Unique properties of the everting principle for locomotion are:
\begin{itemize}
    \item The system is not stationary in itself but changes constantly, which means that the part of the tube being at the front changes while everting. Thus, it is more challenging to mount something (e.g. sensors) to the tip on the outside of the tube. Our solution to the head mounting is described in \Cref{sec:sensors}.
    \item The tube is soft and deformable, which renders the analysis, control, and simulation of the locomotion more challenging. Our solution to controlled lateral steering is described in \Cref{sec:lateral_movement}.
    \item Fully controlled backward movements of the tube by “everting inwards” leads to kinks of the tube, a modeling and controls challenge we will address with future designs. 
\end{itemize}
Despite the challenges associated with these properties, this approach to locomotion is still the most promising when it comes to achieving operation ranges of more than \SI{10}{\metre} and high maneuverability.

\subsection{Supply Box}

\begin{figure}
    \centering
    \includegraphics[width=\linewidth]{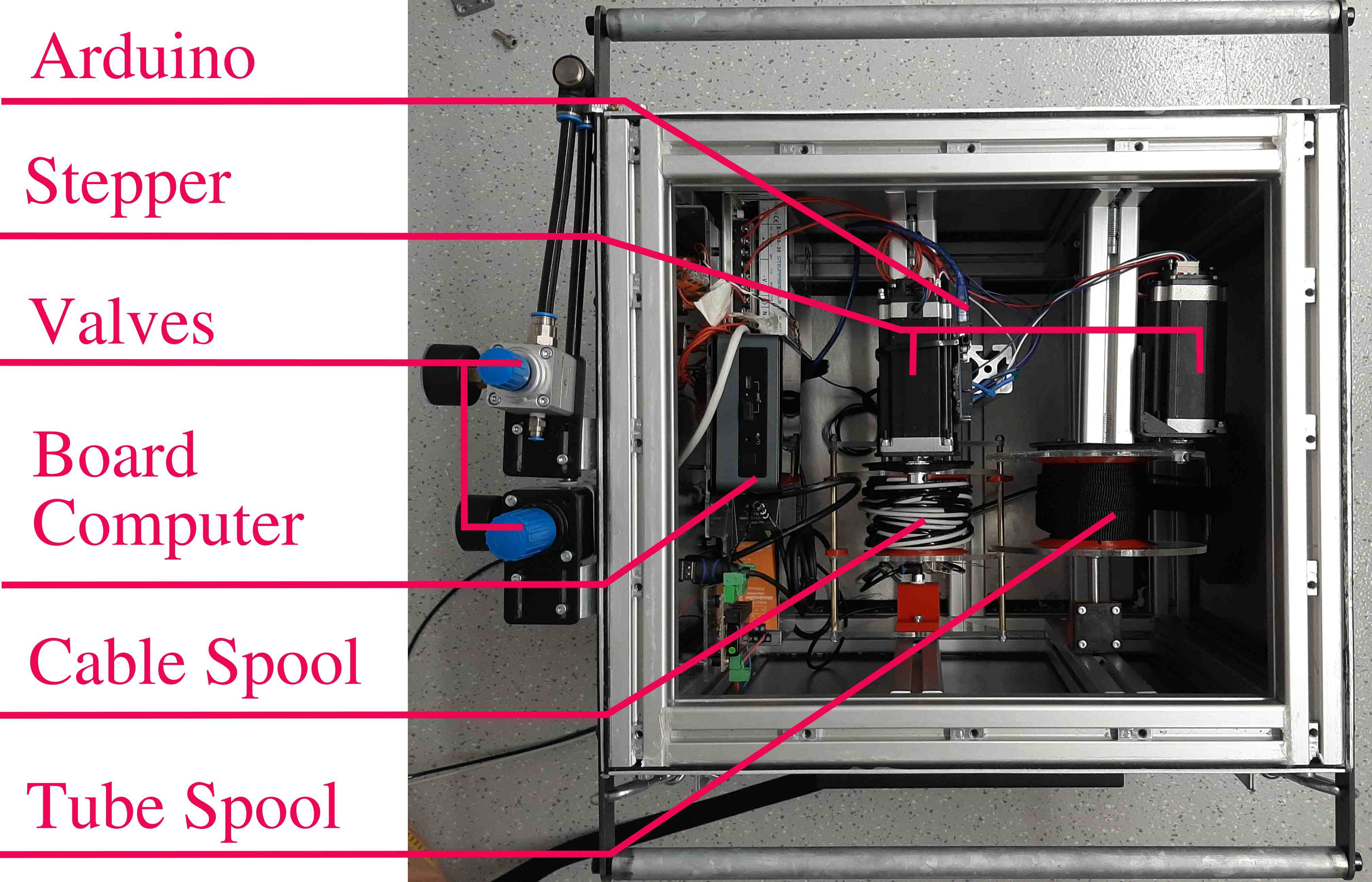}
    \caption{Supply box with removed lid.}
    \label{fig:box}
\end{figure}
The tube is stored in a supply box (shown in Fig.\,\ref{fig:box}), which also houses the cable (see \Cref{sec:lateral_movement}) and the onboard computer. The box is pressurized up to \SI{20}{\kilo\pascal} relative to ambient air pressure to inflate the tube. This box will always remain outside of the debris thus its weight and dimensions are not a limiting factor for the movement of the robot, but for convenient operation it was aimed for a weight which two adults can carry. The weight of the box ended up to be around \SI{50}{\kilo\gram}. The majority of the weight stems from the aluminium plates with \SI{27}{\kg}, aluminium profiles weight \SI{10.5}{\kg} and electronic parts \SI{6}{\kg}. The remaining weight is from screws, sealant, tube material, buttons and air valves. The rescue troops are equipped with pneumatic tools such as jack hammers and have therefore a compressor with included generator on site. This simplified the design requirements, since the robot can rely on existing equipment. In the field the robot's theoretical maximum power draw of \SI{350}{\watt} can be provided by the current equipment.
The tube and cable are stored on two spools which are driven by separate Nema 23 stepper motors with \SI{3}{\newton\meter} torque. The spools cannot be driven by the same motor without a transmission because the supply cable moves at the velocity of the head, which is only half of the speed of the tube. Slip rings are necessary between the stationary supply box and the rotating spools. Aluminium profiles build a cubic structure. The grooves in the profile are used to mount all axles and electronic parts inside the box. The aluminium plates sealing the box from air leakage are also mounted to them.

\subsection{Lateral Movement}
\label{sec:lateral_movement}

\begin{figure}
    \centering
    \includegraphics[width=0.48\textwidth]{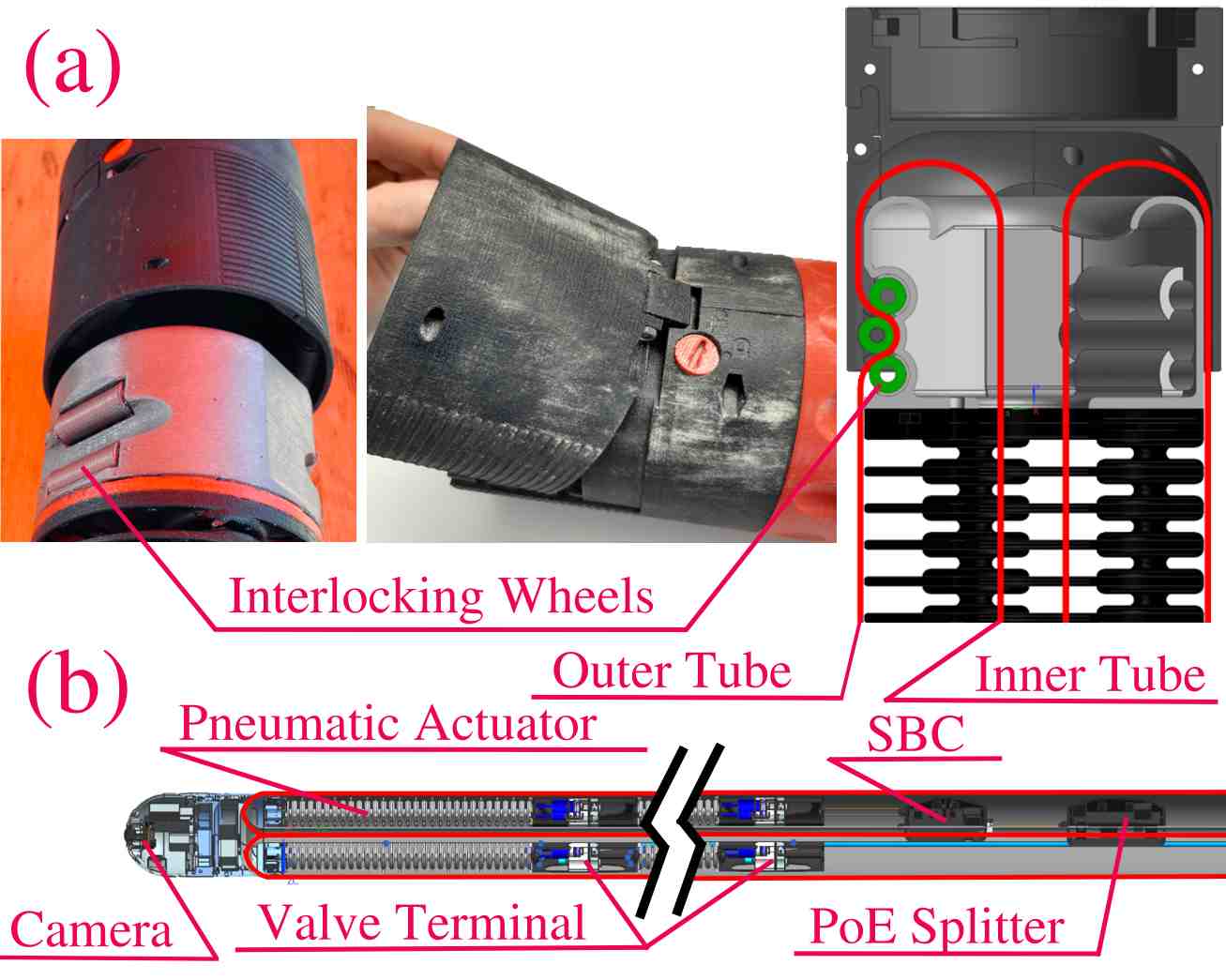}
    \caption{(a) Two exterior views and a cross-sectional view of the interlocking mechanism between the head and the tube. (b) Cross section of the internal robot within the tube is shown. Note that the second segment of the pneumatic actuator has been omitted for simplicity.}
    \label{fig:tube}
\end{figure}

To control the forward locomotion of the tip of the everting tube, we propose a structure that can actively be bent at the front inside of the tube. This structure is henceforth referred to as the \textit{internal robot}. An overview of the system is shown in Fig.\,\ref{fig:IR}. The internal robot mainly consists of actuator segments and valve terminals. The segments are made of 3D printed nylon (FS3200PA-F) and allow the robot to deflect laterally when pressurized with up to \SI{400}{\kilo\pascal}. The valve terminals regulate the airflow from and to the soft actuators. Each terminal consists of proportional flow valves and a custom \ac{pcb} with a micro controller. To minimize the volume of air of which the pressure has to be controlled, the valve terminals are located directly at their associated actuator respectively. The internal robot requires pressurized air and a communication interface. To achieve this, the internal robot is connected to the supply box with a slim air hose as well as a single \ac{poe} cable, combining Ethernet and power lines. To split the data and power and manage communication between the low-level microcontrollers, the head and the computer in the supply box, the robot carries a \ac{poe} splitter as well as an \ac{sbc} packaged in two small cases.

%
\subsection{Sensors}
\label{sec:sensors}

The ability to steer the robot remotely without direct line of sight makes a good sensing payload mandatory. The head, which contains all necessary sensors, is located at the tip of the robot. It is attached to the internal robot with an interlocking mechanism similar to \cite{Coad_2020}. Fig.\,\ref{fig:tube}a shows Roboa's interlocking mechanism. The interlocking wheels of the head and the ones on the internal robot slightly interlock in each other to prevent the head from detaching. A small gap between the wheels allow the tube to slip through. All interlocking wheels are supported by small ball bearings to minimize the friction. The head contains a greyscale 640x480 pixel camera and an \ac{imu} for navigation and a speaker and microphone to communicate with located victims. The head is attached to the internal robot in a such a way that they always share the same orientation. Therefore, steering to the left for the operator on the screen corresponds to the same motion of the robot in the field. Since the head is isolated and unreachable by a cable (see Fig.\,\ref{fig:tube}) -- due to the everting tube between the internal robot and the head -- it is battery powered and transmits its readings to the internal robot wirelessly. The battery of the head lasts for approximately one hour. The on-board \ac{sbc} of the internal robot then transmits the data back to the operator. Due to the small distance between the head and the receiver, the connection proved to be reliable even when operating at the maximum range.

\subsection{Control}

For the control of the lateral motion the robot has a draw-wire sensor on each of the three pneumatic actuators to calculate the deflection by measuring the elongation along three directions parallel to the actuator. The measured values are compared against desired reference values which are computed using a piecewise constant curvature model\,\cite{webster2010design}. The measured deviation is then used by a PID controller to adjust the pressure in the corresponding actuator strand. The reference values for the actuator deflection are mapped from inputs given using a conventional joystick. The mapping prioritizes movements in the first, frontal segment before actuating the second one, since the bending of the first actuator is usually more desirable for the operator as observed in the field. Additionally, the second actuator has to move the first one, which can be difficult because of the added weight. The unique principle of deflection among vine robots allows precise angular control of the deflection of the tip, which is unprecedented for vine robots in this size.
The throttle lever of the joystick corresponds to forward motion of the tube. Forward motion is caused by the two stepper motors, which are controlled by an Arduino. Currently, the pressure in the box is only controlled manually by a proportional valve.

%% file: 03_results.tex
\section{Experimental Evaluation}
\label{sec:results}

\begin{figure}[hbp]
   \centering
   \includegraphics[width=0.48\textwidth]{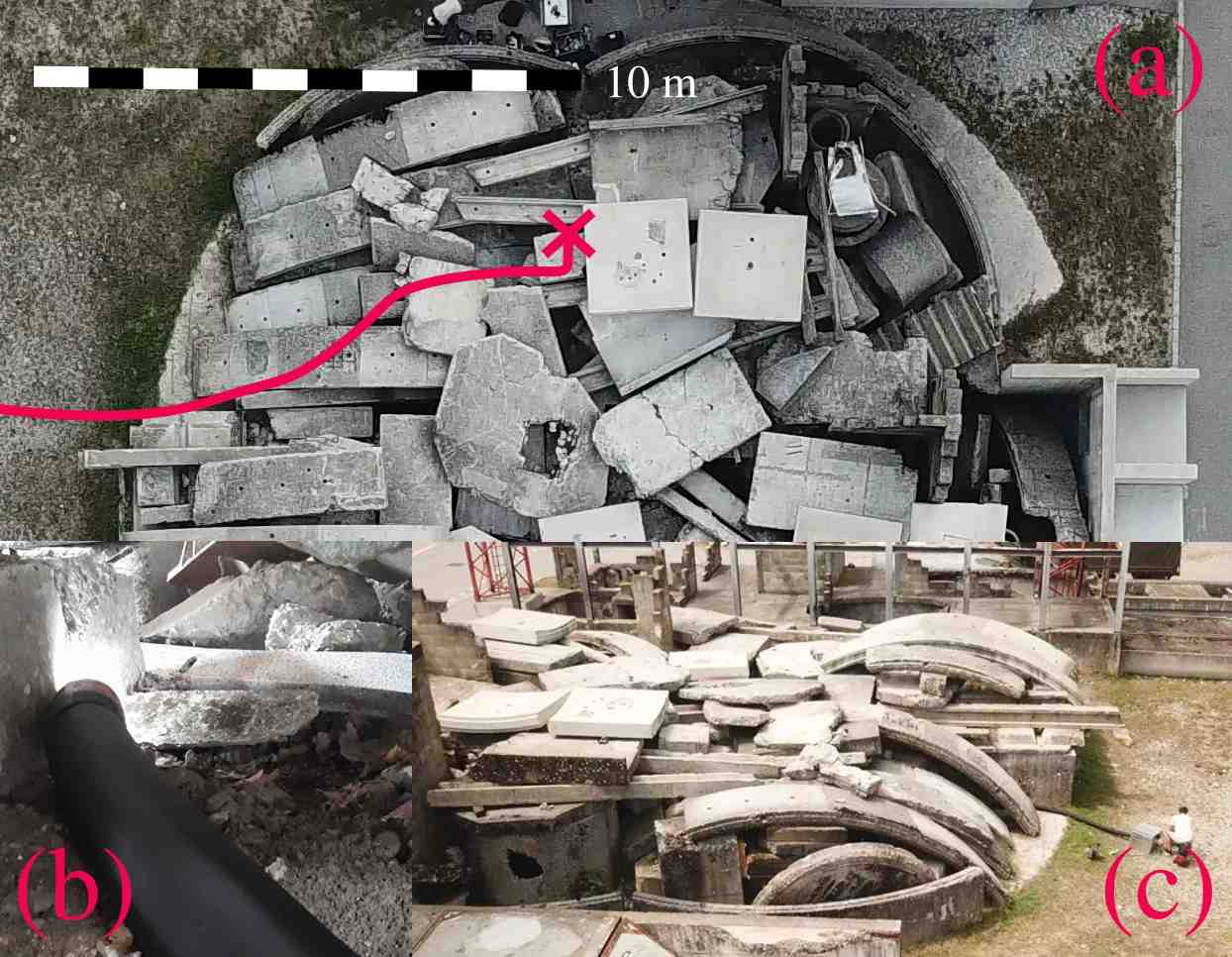}
   \caption{(a) Followed path (approximately) to find a team member. The time to reach the goal was on average \SI{20}{\minute}.
        (b) Picture of the robot inside the debris. 
        (c) Debris in which RoBoa was tested. On the right side of the image the robot, as well as the operator.}
   \label{fig:debris}
    \end{figure}



%

\subsection{Real-world Evaluation}
To test RoBoa's capabilities under realistic conditions, RoBoa was taken to the Swiss Rescue Troop's training site in Wangen an der Aare. The debris in which RoBoa was tested and the approximately followed path are shown in Fig.\,\ref{fig:debris}. The goal of this test was to find a person hidden inside of the debris by remote-controlling RoBoa and getting feedback solely through the camera feed. On its way into the debris, RoBoa's pilot had to maneuver over obstacles and through openings too small for a person. Eventually, the robot found the hidden person and was able to communicate with her.

The test on the training ground proved the operational capability of RoBoa in a realistic environment. It was able to follow a \SI{10}{\metre} long path with several obstacles, two curves and one 90$^\circ$ turn without getting stuck inside the debris.
This test was performed by a pilot teleoperating the robot at the ground station with the provided camera feed. The results of this first test in a realistic environment were very positively received by the Swiss Rescue Troops.

\subsection{Characterization of Capabilities}

The robot uses the everting principle and moves forward at a maximum velocity of \SI{6}{\metre\per\minute} depending on the environment. Backward movement is not possible in the current state. It is a much greater challenge than expected at first sight because of the kinking characteristic of the tube. 
When RoBoa is fully deployed, it has a length of \SI{17}{\metre}. The length can be adjusted by using a longer or shorter tube. The tube can be used to pave a way to the victim and to pull the victim out of the debris. However, the tube has to be replaced after four to five deployments because it can get damaged in the debris and the compressor can only compensate for a certain amount of leakage.
The part of the robot with the largest diameter is the head with a diameter of \SI{101}{\milli\metre}.
Lateral movement is possible by up to 120$^\circ$ left and right. Due to the weight of the head attachment and the cylindrical shape of the internal robot it is able to bend upwards by 50$^\circ$ when supported by the environment. Without support it is able to lift its head about \SI{20}{\centi\metre} from the ground achieving a bending angle of 20$^\circ$ before it looses its balance and rolls over due to its circular shape. At high angles the internal robot obstructs the eversion process which prohibits simultaneous forwards motion with large deflections. Instead, large deflections can be used to look around before straightening the actuators again to continue moving forwards. In general, the maneuverability of RoBoa is limited in open spaces because the part of the tube that is behind the internal robot cannot hold shapes by itself and needs external structures to remain bent. Only turns of about \SI{15}{\degree/\metre} are possible on frictional surfaces such as grass or gravel.
RoBoa's head provides the pilot with a camera feed which makes the steering using a joystick very intuitive. Its built in speaker and microphone allows the pilot to communicate with the victim inside the debris. In addition, hardware components for precise localization are implemented, but the necessary software does not run yet. However, sensing in the head is easily adjustable independent from the rest of the robot. Setting up the system is not possible for a person that has not been instructed.

In summary, RoBoa fulfills most of the set requirements in regards to locomotion, dimension, steering and communication. However, there is still potential to improve the robot (e.g. backward movement, localization).

%% file: 04_conclusion.tex
\section{Conclusion}

\subsection{Discussion}

The key capabilities of the realized prototype are a length at maximum tested extension of up to \SI{17}{\metre}, a maximum diameter of \SI{101}{\milli\metre}, and a maneuverability of up to 120$^\circ$ bending angle and down to \SI{20}{\centi\metre} turning radius. This steerable everting tube robot can reach hard to access areas many meters into a field of rubble, making it particularly suitable for \ac{sar} applications in confined and unstructured environments.

Simplified localization is enabled by the positioning of the internal robot and by the tube connection from the supply box to the robot's head.
The camera, microphone and speakers allow steering the robot and communicating with victims without seeing the whole robot.

Besides the actual deployment as a \ac{sar} robot in test scenarios, the novelty of the presented design lies in 
the combination of the everting vine robot concept for locomotion with a multi-segment, pneumatic actuator for steering.
This decouples locomotion from lateral movement:
The vine robot principle allows for more versatile forward locomotion in debris fields than other locomotion approaches because it minimizes friction.
The internal robot offers advantages for navigating in narrow environments. Its attached valve terminals enable a portable and decentralized pressure supply.

Compared to a body-steered vine robot,
our actuated internal robot allows for a high bending angle and a low turning radius, can lift high payloads at the tip of more than \SI{1}{\kilogram}, and enables s-shapes.
As it is always located at the front, the unused actuator-length is minimized and 
the robot only steers where necessary.
This is particularly useful for the debris environment's unique nature with many constraints due to obstacles and small spaces. The steerable front decides on the path and the rest of the robot follows due to the constraints. 
However, this means that the robot requires interaction with the environment to steer successfully and cannot direct its motion properly in open spaces.

Using feedback and state estimation at the internal robot, precise steering at the tip can be achieved.

The decoupling enables the development of such a long vine robot as the tube does not need to contain any actuators itself. This simplifies both tube manufacturing and everting. Furthermore, the tube can be easily replaced if damaged during deployment.

Nevertheless, the decoupling also comes with some drawbacks. The usage of the internal robot adds complexity to manufacturing, friction inside and weight to the system. The last point makes it more difficult to overcome longer cracks in comparison to a body-steered vine robot. Moreover, the rigid structure of the head and the internal robot prohibits the vine robot's natural capability to shrink.

To sum up, the prototype showed that this concept has the potential to successfully serve as a \ac{sar} robot that can navigate through debris and locate trapped victims. The feedback received during the evaluation of the system together with the Swiss Rescue Troops at their training site was that RoBoa would provide an immediate utility for their operations.

\subsection{Future Work}

The demonstrated utility of the system led to a follow-up project funded by the Swiss Drone and Robotics Centre. To soon support rescue workers in day to day training and active application of the system, the goal is to provide them with a device at technology readiness level 7.

The focus lies on increasing the system's reliability and capability and taking the current subsystems from a prototype stage to a reproducible level, particularly the internal robot and the supply box.
To do so, we would like to explore alternative cross-sections for the internal robot to overcome problems with moving upwards in certain situations.
The weight the internal robot adds to the system is intended to be reduced, for example by designing a lighter valve terminal.
Special 3D-printing technologies will minimize friction in the everting mechanism and a new interconnection system will increase the internal robot's modularity and thus will simplify its application.
A smaller and lighter supply box will enable faster deployment of the robot.
Furthermore, although the tube material is already resistant against adverse environmental conditions, its life-time is limited. We would like to investigate other materials and ways to further improve this.



New functionalities that we are planning to add are backward movement of the tip, a water supply line for victims, and a more capable sensor system to localize and identify victims, for example \ac{slam} and chemical noses.